# Enabling SQL-based Training Data Debugging for Federated Learning

Yejia Liu*
Simon Fraser University
Burnaby, BC, Canada
yejial@sfu.ca

Weiyuan Wu*
Simon Fraser University
Burnaby, BC, Canada
youngw@sfu.ca

Lampros Flokas
Columbia University
New York, NY
lamflokas@cs.columbia.edu

Jiannan Wang
Simon Fraser University
Burnaby, BC, Canada
jnwang@sfu.ca

Eugene Wu
Columbia University
New York, NY
ewu@cs.columbia.edu

## ABSTRACT

How can we debug a logistical regression model in a federated learning setting when seeing the model behave unexpectedly (e.g., the model rejects all high-income customers' loan applications)? The SQL-based training data debugging framework has proved effective to fix this kind of issue in a *non*-federated learning setting. Given an unexpected query result over model predictions, this framework automatically removes the label errors from training data such that the unexpected behavior disappears in the retrained model. In this paper, we enable this powerful framework for federated learning. The key challenge is how to develop a security protocol for federated debugging which is proved to be secure, efficient, and accurate. Achieving this goal requires us to investigate how to seamlessly integrate the techniques from multiple fields (Databases, Machine Learning, and Cybersecurity). We first propose FEDRAIN, which extends RAIN, the state-of-the-art SQL-based training data debugging framework, to our federated learning setting. We address several technical challenges to make FEDRAIN work and analyze its security guarantee and time complexity. The analysis results show that FEDRAIN falls short in terms of both efficiency and security. To overcome these limitations, we redesign our security protocol and propose FROG, a novel SQL-based training data debugging framework tailored for federated learning. Our theoretical analysis shows that FROG is more secure, more accurate, and more efficient than FEDRAIN. We conduct extensive experiments using several real-world datasets and a case study. The experimental results are consistent with our theoretical analysis and validate the effectiveness of FROG in practice.

## 1 INTRODUCTION

Companies and organizations increasingly need to balance the desire to share and augment their training data with additional data attributes in order to improve ML model quality with the challenge of keeping private data secure. Recent work in federated learning leverages secure multi-party computation to develop distributed training and inference protocols that enable organizations to collaboratively train models over their joined data *without* explicitly sharing their private data with each other nor a third-party.

Training data quality is critical to developing accurate and unbiased ML models [31, 51]. However, in federated learning, training data errors can come from any of the data sources, and we anticipate

*The first two authors contributed equally to this research.

Table 1: Debugging training data for federated learning.

| Method | Complaint Type | Secure? |
|---|---|---|
| Model loss | Not Supported | ✓ |
| Influence function [32] | Instance-based | ✗ |
| Rain [59] | SQL-based | ✗ |
| **Our work** | **SQL-based** | ✓ |

the need for efficient and effective federated data debugging techniques. Specifically, when a biased or inaccurate federated learning model mispredicts in a way that affects downstream analysis results, can we automatically identify the training examples that most contributed to the downstream error in a way that retains federated learning's security guarantees? Without solving this problem, companies will be cautious about deploying federated learning models in production.

*Example 1.1 (Federated ML Debugging).* Figure 1 borrows a real-world Financial use case from Yang et al. [3]. Fintech Company A wants to work with a Bank B to build a risk assessment model that approves customer credit card applications. For simplicity, Company A's data has schema (ID, Income, Label), and joins on ID with Bank B's data with schema (ID, Deposit). They first securely train model $M(\text{Income}, \text{Deposit}) \rightarrow \text{Label}$ over the joined training data $D^A \bowtie_{ID} D^B$. Later, they securely make inferences on the joined inference data $I^A \bowtie_{ID} I^B$ and materialize the predictions in table P with schema (ID, Label). Company A (and Bank B) may now use the table $P$ in its application and downstream analytics. For instance, a data scientist Lucy at Company A builds a dashboard that visualizes the percentage of rejections (label=0) from wealthy applicants:

```
SELECT COUNT(·)/Total_Count AS ratio
FROM P ⋈_ID I^A
WHERE I_A.Income > 200,000 AND P.Label = 0
```

If the result is high (say, 0.5), then Company A is rejecting half of its the high-income applicants. If, after checking the inference dataset, Lucy does not find any errors, then the issue may be due to errors in the training data. Ideally, Lucy should be able to specify that "this ratio should actually be zero" (we call this statement a *complaint*), and be presented with the training record IDs that, if deleted from either party's training data, would resolve the complaint. Further, this debugging process should maintain the same security properties as training and inference. Otherwise, Company A and Bank B are unlikely to deploy such a model in production.

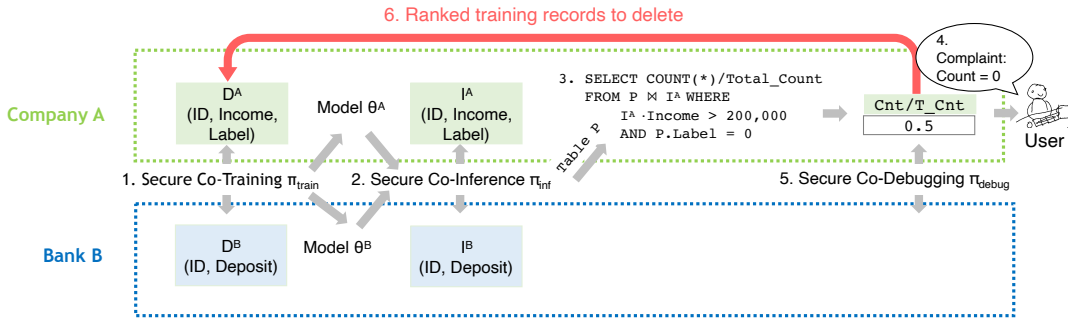

Figure 1: Overview of this paper's Federated training and debugging workflow, focusing on Company A's actions (the process is the symmetric for Bank B). Company A issues a COUNT query and then complains about the returned queried result. Based on the complaint, the debugging framework would generate a list of training data deletion IDs that most resolve the complaint.

**Possible Solutions.** Despite the popularity of inference queries that combine relational operators and model inference, there are few techniques today that identify erroneous training records that directly affect errors in the query results. Table 1 summarizes the four primary approaches today. They adopt an intervention-based debugging framework [39] and aim to remove a subset of training data such that if removed, and the model was re-trained, the new model would not lead to the unexpected query answer.

(1) **Model Loss** is a simple baseline that iteratively removes training examples from the highest to lowest training loss until the complaint is resolved. The training losses can be trivially obtained during federated inference, thus this method is secure. However, model loss is independent of the complaint, and so this will likely remove many irrelevant training records.

(2) **Influence Function** partially overcomes the preceding limitation. It allows the user to specify an instance-based complaint, i.e., individual model mispredictions, and quickly approximates the influence of removing each training example on the misprediction [32]. For the above example, Company A can point to a specific high-income customer and ask why this customer's credit card application is rejected by the model. This method iteratively ranks and removes training records based on their estimated influence on the complaint until the complaint is resolved. While this improves over the baseline, it is limited to misprediction labels and not downstream analytics, and is not secure in a federated setting.

(3) **Rain** [59] generalizes complaints to SQL-based queries, i.e., complain about an unexpected query answer, and quickly approximates the influence of removing each training example on query result. It does so by transforming the training and query workflow into an end-to-end differentiable function. This generalizes the previous method, as instance-based complaints are a degenerate query. For the above example, Company A can directly complain that why the ratio is too high and it should be zero. Rain iteratively ranks and removes training records that most increased the ratio until the complaint is resolved. Unfortunately, Rain is not secure.

(4) **This paper** presents a secure debugging framework that provides rich Rain-style complaints but uses an efficient secure computation protocol to avoid leaking private data.

**Scope of this paper.** This work presents the first SQL complaint-based training data debugging approach for federated learning that is secure and efficient. We identify a setting that is common and amenable to efficient protocols. i) Logistic Regression Models are one of the most common classification models. Some of the first federated learning algorithms were designed for logistic regression [23] and have been implemented in popular open-source federated learning libraries [2]. ii) Two Party: most existing federated learning algorithms have studied a third-party setting [15, 23, 60] that relies on a third party to facilitate collaboration between the cooperating parties. However, this requires identifying such a trusted party and is an additional risk for data leakage. In contrast, our work targets the more practical but challenging setting that removes the dependency on the trusted third party. iii) Vertical Federated Learning is where the data is vertically partitioned across the parties. This is more complex as it requires decomposition of loss functions into each party [15, 23, 60].

**FedRain and Frog.** Our first solution FedRain directly extends Rain to the federated setting. FedRain follows the same computation procedure as Rain. For each step in Rain that is not secure, FedRain uses homomorphic encryption to encrypt the private data and then compute over encrypted data. Since not all the computations can be applied to encrypted data, we need to carefully design secure protocols to e.g., calculate the gradient of an SQL query and compute a *hessian-vector-product*. We also present the security and complexity analysis.

Federated learning is bottlenecked by data encryption and rounds of communication, and the FedRain protocol does not scale to large training data sizes. Furthermore, FedRain needs to limit the number of stochastic gradient descent iterations to less than the number of features in order to ensure security. However, this is typically far lower than the iterations needed for logistic regression to converge. Thus, FedRain often cannot reach high model accuracy without breaking the security guarantee.

To overcome these limitations, Frog is a secure, efficient, and accurate training data debugging framework tailored for federated learning that makes two main technical contributions. First, we decompose the logistic regression model structure so that each party can train a local model and then securely combine the model parameters. This lets each party efficiently train their local models, but still exploit features from the other party to build an accurate model. Second, we design efficient and secure protocols for model training and debugging in a federated learning setting. We prove that the protocol is secure and has much lower time complexity than FedRain.

We conduct extensive experiments to evaluate FedRain and Frog, and compare them with the baselines above. Our empirical results are consistent with our theoretical analysis showing that Frog is more accurate, more efficient, and more secure than FedRain. We also present a case study to demonstrate Frog can effectively resolve a real SQL-based complaint. In the case study, the model is to predicate whether an employee should get a high salary. The query is to compute the percentage of predicated high-salary female and male employees, respectively. The user complains that "the difference between the two percentage values should actually be zero". That is, the user wants the model to make a fair prediction. Frog is able to help the user to reduce the difference by a large margin with very little loss in model's F1 Score.

**Contributions.** The following summarizes our contributions:

- We are the first to study how to enable SQL-based training data debugging for federated learning. We formally define the problem and discuss the limitations of baseline solutions.
- We propose FedRain, which extends Rain to the federated learning setting. We address several technical challenges in the design of its security protocol and prove its security guarantee and analyze its time complexity.
- We propose Frog, a novel federated debugging framework. We design a training and a debugging security protocols for Frog and prove their security guarantees and analyze their time complexities.
- We conduct extensive experiments using real-world datasets and a case study. The results that i) FedRain and Frog enable SQL-based training data debugging for federated learning; ii) Frog significantly outperforms FedRain in terms of both efficiency and accuracy; iii) Frog can effectively resolve a real SQL-based complaint.

Next, we will present the problem definition and introduce the necessary background for Rain and homomorphic encryption. Section 4 and 5 will present FedRain and Frog respectively, and Section 6 presents our evaluation and case study.

## 2 PROBLEM DEFINITION

In this section, we define our federated debugging problem. Before that, we first introduce some background knowledge about logistic regression, security model, and federated training and inference.

### 2.1 Background

**Logistic Regression.** Logistic regression is a commonly used classification model in machine learning. For a feature vector $x$, the logistic regression model predicts $x$ of having a class label of 1 as

$$h_\theta(x) = \frac{1}{1+e^{-\theta^\mathsf{T} x}}, \quad (1)$$

where $\theta$ is the model parameter. The parameter $\theta$ is learned by maximizing the log likelihood on a training dataset $D$. Specifically, given a training dataset $D = \{(x_i, y_i)\}_{i=1}^n$, where $x_i$ represents a feature vector and $y_i \in \{0, 1\}$ is the class label of $x_i$, the training algorithm aims to find the best $\theta$ to maximize the following

likelihood:
$$\prod_{i=1}^n \left(h_\theta(x_i)^{y_i}(1-h_\theta(x_i)^{1-y_i})\right)$$

This is equivalent to finding the best $\theta$ to minimize the following log likelihood:

$$\begin{aligned} L(\theta) = & -\frac{1}{n}\sum_{i=1}^n \ell_i(\theta) \\ & -\frac{1}{n}\sum_{i=1}^n \left(y_i \log h_\theta(x_i) + (1-y_i)\log(1-h_\theta(x_i))\right). \end{aligned} \quad (2)$$

In a federated learning setting, the training dataset $D$ is vertically partitioned as $D^A$ and $D^B$, stored in *Party A* and *Party B*, respectively. We assume that *Party A* contains a subset of features and the class label, $D^A = \{(x_i^A, y_i)\}_{i=1}^n$, *Party B* contains the other subset of features, $D^B = \{(x_i^B)\}_{i=1}^n$. The number of features contained in $D^A$ and $D^B$ are denoted as $m^A$ and $m^B$. For ease of presentation, we assume that $D^A$ and $D^B$ share common unique IDs. We denote their join result by $D = D^A \bowtie_{ID} D^B$. This set intersection operation can be done in a privacy-preserving manner [13, 45].

To avoid data leakage, *Party A* and *Party B* have to follow a *security protocol* for training, inference, and debugging, respectively.

**Security Model.** We use a common security model in federated learning, named *honest-but-curious* [15, 36, 60]. That is, Party A and Party B will always follow the specified security protocol. However, they may also try to learn the protected data from another party given the messages received.

We consider the protocols that two *honest-but-curious* parties *Party A* and *Party B* collaboratively compute a function by exchanging messages to each other in multiple rounds. Formally, given a function $\mathcal{F}$, *Party A* and *Party B* run a protocol $\pi_\mathcal{F}$ to compute $\pi_\mathcal{F}(I^A, I^B)$, where $(I^A, I^B)$ is the input of the function, $I^A$ is from *Party A*, and $I^B$ is from *Party B*. Let $(O^A, O^B)$ denote the output of the function, i.e., $(O^A, O^B) = \pi_\mathcal{F}(I^A, I^B)$. *Party A* has access to $I^A$ and $O^A$. *Party B* does not want *Party A* to guess out what $I^B$ and $O^B$ are. Therefore, we say a protocol is **secure** against *Party A* if there are infinite number of $(I^{B'}, O^{B'})$ such that $(O^A, O^{B'}) = \pi_\mathcal{F}(I^A, I^{B'})$, and vice versa for *Party B*. This is a common security definition adopted by many existing works [17, 27, 35, 44, 63].

*Definition 2.1 (Two-Party Security Protocol).* Given a function $\mathcal{F}$, two *honest-but-curious* parties, an input $I^A$ from *Party A*, and an input $I^B$ from *Party B*, a two-party security protocol, denoted by $\pi_\mathcal{F}$, is a multiple round message passing protocol between *Party A* and *Party B* that compute $\pi_\mathcal{F}(I^A, I^B)$ and derive $(O^A, O^B)$. The protocol contains a sequence of messages $\{M_i^{A \to B}, M_i^{B \to A} \mid 1 \le i \le N\}$ that *Party A* and *Party B* will send to each other for each round $i \in [1, N]$.

**Federated Training.** For federated training, the inputs are $D^A$ and $D^B$, and the function $\mathcal{F}$ is $\min_\theta L(\theta)$ defined in Equation 2, where each training example $(x_i, y_i)$ comes from $D = D^A \bowtie_{ID} D^B$, and the outputs are $\theta^A$ and $\theta^B$, where the model parameter $\theta$ is the concatenation of $\theta^A$ and $\theta^B$, denoted by $\theta = \theta^A | \theta^B$.

In order to train a logistic regression model while keep data secure, *Party A* and *Party B* should follow a protocol $\pi_{train}$ which is defined as:

*Definition 2.2 (Federated Training).* Given two training datasets, $D^A$ and $D^B$, the goal of federated training is to design a two-party security protocol, denoted by $\pi_{\text{train}}$, that trains a logistic regression model over $D^A \bowtie_{ID} D^B$ which computes $\min_\theta L(\theta)$.

**Federated Inference.** Once a model is trained, it will be applied to an inference dataset $I = I^A \bowtie_{ID} I^B$ to do model inference. $I_A$ and $I_B$ are the inference datasets owned by *Party A* and *Party B*, respectively.

For federated inference, the inputs are $I^A$, $\theta^A$, $I^B$, and $\theta^B$, the function $\mathcal{F}$ is $h_\theta(x)$ defined in Equation 1, where each $x$ is from $I^A \bowtie_{ID} I^B$, and the output is the predicated class label of each $x$.

*Definition 2.3 (Federated Inference).* Given two inference datasets $I^A$ and $I^B$, and the model parameter $\theta^A | \theta^B$, the goal of federated inference is to design a two-party security protocol, denoted by $\pi_{\text{inf}}$, that does the logistic regression inference over $I^A \bowtie_{ID} I^B$ which computes $h_\theta(x)$ for each $x$.

## 2.2 Our Problem: Federated Debugging

Once a model is trained, each party can securely make inferences on the joined inference data and materialize the predictions in table P with schema (ID, Label). Each party can issue aggregation queries on the joined result between table P and its own inference data. Without loss of generality, suppose that the queries are executed in *Party A*. In the following, we first present the supported inference queries, and then define our federated debugging problem.

**Inference Query.** We allow *Party A* to issue an aggregation query on its inference data $I^A$ and the table $P$ in the following form:

```
SELECT agg(·) FROM P ⋈_ID I^A
WHERE C_1 AND C_2 AND ··· C_m
GROUP BY G_1, G_2, ···, G_k
```

The *agg* function can be **count, avg** or **sum**. $I^A$ is the inference dataset held by *Party A*. $C_i$ ($i \in [1, m]$) is a filter condition and $G_i$ ($i \in [1, k]$) is a group-by attribute. Like Rain [59], P.Label can appear in the SELECTION, WHERE, or GROUP BY clause. For example, the query in Figure 1 puts P.Label = 0 in the WHERE clause. We denote this query result as $Q(I^A; \theta)$, where $\theta$ represents the model that populates the table $P$.

**Federated Debugging.** We call the statement that whether $Q(I^A; \theta)$ satisfies an expected value a *Complaint*. Formally, we define *Complaint* as follows:

*Definition 2.4 (Complaint).* A complaint $c(\cdot)$ is expressed as a boolean constraint over the query result $Q(I^A; \theta)$, where $op \in \{=, \leq, \geq\}$ and $v$ may take any value in the aggregation result's domain.

$$c(Q(I; \theta)) = \begin{cases} \text{True,} & \text{if } Q(I; \theta) \text{ op } v \\ \text{False,} & \text{otherwise} \end{cases} \quad (3)$$

For the complaint in Figure 1, the expected value is $v = 0$, and the complaint $c(\cdot)$ is to check whether the query result $Q(I^A; \theta)$ is equal to 0. For the model $\theta$, the query result is $Q(I^A; \theta) = 0.5$, thus it violates the complaint (i.e., $c(Q(I^A; \theta))$ = False). Our goal is to identify the minimum number of training examples such that if they were removed, and the model was retrained, the updated model $\theta'$ would lead to a new query result $Q(I^A; \theta')$ that satisfies the complaint (i.e., $c(Q(I^A; \theta'))$ = True). Definition 2.5 formally defines the problem.

*Definition 2.5 (Federated Debugging).* Given two training datasets, $D^A$ and $D^B$, the model parameter $\theta^A | \theta^B$, an inference query $Q(I^A; \theta)$, and a complaint $c(\cdot)$, the goal of federated debugging is to design a two-party security protocol, $\pi_{\text{debug}}$, that searches for a minimal sized $\Delta$ such that if it were removed, the complaint would be resolved. More formally, *Party A* and *Party B* aim to compute the following function:

$$\min_{\Delta \subseteq D^A \bowtie_{ID} D^B} |\Delta|$$
$$\text{s.t.} \quad c(Q(I^A; \theta')) = \text{True}$$
$$\text{where} \quad \theta' = \arg\max_\theta L(\theta)$$

**Challenges.** We face several challenges to solve this problem. Firstly, this problem touches multiple fields (Databases, Machine Learning, and Cybersecurity). It requires us to investigate how to seamlessly integrate the techniques from these fields. Secondly, there are three aspects to evaluate a security protocol: accuracy, efficiency, and security. It is not easy to design a security protocol that performs well in all three aspects. Thirdly, we need to not only provide theoretical guarantees but also implement our approach in order to gain a deep understanding of its performance.

## 3 PRELIMINARIES

Our framework is built on the Rain debugging framework, as well as the Paillier Encryption Scheme for security guarantee.

### 3.1 Rain [59]

Rain is a SQL-based training data debugging framework proposed in a non-federated learning setting. In detail, Rain assumes that there exists an ML model which is trained on the training data $D$ and then its prediction (i.e., M.predict) is embedded in a SQL query. After the query is executed, a user can issue a complaint on the query result. Rain then takes the complaint as input and produces a ranked list of the training examples in $D$ based on how much each training example contributes to the complaint. Rain proposes an iterative debugging framework that takes a budget $K$ as input and runs the following steps until the budget is exhausted:

(1) Generate the ranked list;
(2) Remove the top-$k$ training examples from the ranked list;
(3) Retrain an ML model on the new training set;
(4) Set $K = K - k$. Repeat (1)-(4) until $K < 0$.

On the technical side, Rain solves two challenges: 1. How to efficiently compute the effect on the query result for deleting each training example? 2. How to make the SQL query differentiable with respect to the model parameters so that continuous optimization techniques can be applied for solving the challenge 1. The solution that Rain comes up with is: Rain first uses provenance polynomial [6, 22] to convert a SQL query into a formula. After that, Rain relaxes the discrete variables in the formula into continuous variables so that the formula becomes differentiable. Lastly, Rain connects the relaxed formula with influence function [32] to compute the score for each training example, which indicates how much it can address the complaint by deleting that training example.

For example, the query in Figure 1, can be expressed into a formula as

$$\sum_{i \in I^A_{\text{Income}>200000}} \frac{\mathbb{1}(\text{P.Label}_i = 0)}{\text{Total\_count}}$$

where $I^A_{\text{Income}>200000}$ is a subset of the inference dataset $I^A$ that only contains the records of Income > 200000. $\mathbb{1}$ is the indicator function which returns 1 (0) if the expression inside is 'True' ('False'). P.Label$_i$ is the prediction result for the $i$-th inference record. Then, Rain relaxes this formula further to make it continuous:

$$Q = \sum_{i \in I^A_{\text{Income}>200000}} \frac{\text{P.Label\_prob}_i}{\text{Total\_count}}$$

in which P.Label_prob$_i$ is the probability that the i-th inference record is predicated as 0.

Given the relaxed formula $Q$, Rain computes $Q'H^{-1}E$ to get the influence score of each training example w.r.t. the complaint [59]. Here, $Q' = -\frac{\partial Q}{\partial \theta}$ is the gradient of the relaxed formula with respect to the model parameter $\theta$, indicating the intention of minimizing $Q$. $H^{-1}$ is the inverse matrix of the Hessian matrix of the model loss on the training dataset $D$, i.e. $\frac{\partial^2 L(\theta)}{\partial \theta^2}^{-1}$. $E$ is the element-wise gradient of the model loss on the training dataset: $E = \{\frac{\partial \ell_i(\theta)}{\partial \theta}\}_{i=1}^n$. Please refer to Equation 2 for the definitions of $\ell_i(\theta)$ and $L(\theta)$.

### 3.2 Homomorphic Encryption

Homomorphic encryption (HE) is a type of encryption scheme that allows computations to be operated on the encrypted data without the requirement to decrypt the data. Additionally, the computation result still remains encrypted. Multiple HE schemes are proposed by the security community. Among them, Paillier encryption scheme [43] is the most widely used [15, 23, 60, 63] in Federated Learning.

Paillier encryption scheme is an additive homomorphic encryption scheme which supports producing the encrypted sum of two encrypted values. Let $[\![\cdot]\!]$ denote the encrypted version of a number. Consider two numbers $u$ and $v$. Paillier encryption scheme provides the following operation:

$$[\![u]\!] + [\![v]\!] = [\![u+v]\!] \quad (4)$$

Based on Equation (4), we can define plain text multiplication as

$$u \cdot [\![v]\!] = \sum_u [\![v]\!] = [\![uv]\!] \quad (5)$$

Note that the value $u$ is not encrypted.

Performing computations on encrypted numbers can incur a huge computational cost. For example, the addition on two encrypted numbers can be two to three magnitudes slower than their unencrypted counterpart [23]. This drawback hints us to design protocols that avoid encrypted computation as much as possible to make the solution tractable.

## 4 FEDRAIN: FEDERATED RAIN

In this section, we discuss how to extend Rain to federated learning. We first introduce the existing two-party logistic regression protocol for training and inference in Section 4.1. After that, we propose our FedRain solution for debugging in Section 4.2. Finally, we provide security analysis in Section 4.3.

### 4.1 Training and Inference

FedRain adopts the existing protocol [63] for training a logistic regression model in the two-party federated logistic regression setting. The training protocol assumes the training data $D^A$ and $D^B$ are already aligned and stored on each party. Additionally, the model parameter $\theta$ is split into two parts $\theta^A$ and $\theta^B$ with size $m^A$ and $m^B$. After that, Party A and Party B run gradient descent (GD) collaboratively by exchanging messages. That is, Party A and Party B first compute the gradient of the logistic loss (Equation (2)) for their own parameters $\frac{\partial L(\theta)}{\partial \theta^A} = -\frac{1}{n}\sum_{i=1}^n (y_i - h_\theta(x_i))x_i^A$ and $\frac{\partial L(\theta)}{\partial \theta^B} = -\frac{1}{n}\sum_{i=1}^n (y_i - h_\theta(x_i))x_i^B$. Then, Party A and Party B update their local parameter $\theta^A$ and $\theta^B$ by $\theta^A := \theta^A - \eta \frac{\partial L(\theta)}{\partial \theta^A}$ and $\theta^B := \theta^B - \eta \frac{\partial L(\theta)}{\partial \theta^B}$. $\eta$ is the learning rate of GD. [63] proposed a way to securely compute the gradients $\frac{\partial L(\theta)}{\partial \theta^A}$ and $\frac{\partial L(\theta)}{\partial \theta^B}$: first Party A ask Party B to transfer $\theta^B x_i^B$ for $x_i^B \in D^B$ in plain text, then Party A computes the residual $y_i - h_\theta(x_i)$ and send the encrypted residual to Party B avoiding Party B knowing the labels $y_i$. Party B then computes the gradient $\frac{\partial L(\theta)}{\partial \theta^B}$ with the encrypted residual and send it to Party A for decryption. To avoid Party A knowing Party B's gradient, Party B will also add noise to $\frac{\partial L(\theta)}{\partial \theta^B}$ and later denoise it upon receiving the decrypted gradient from Party A. Throughout the whole process, the training data $D^A$ and $D^B$ as well as the model parameters $\theta^A$ and $\theta^B$ are stored locally in Party A and Party B and remains unknown to the other party. The only message that is transferred in plain text is $(\theta^B)^\top x_i^B$ from Party B to Party A.

On the other hand, the inference stage is straightforward: Party B send $(\theta^B)^\top x_i^B$ for $x_i^B \in I^B$ to Party A and then Party A can directly compute the prediction as $h_\theta(x_i) = \frac{1}{1+e^{-(\theta^A)^\top x_i^A - (\theta^B)^\top x_i^B}}$

### 4.2 Debugging

As mentioned in the preliminaries, Rain computes the $Q'H^{-1}E$ formula to get the influence score for each training sample. In order to extend Rain to federated learning, the key is to design a security protocol to compute $Q'H^{-1}E$, i.e. the influence score of removing each training example on the query result.

#### 4.2.1 Compute $Q'$.
We will show the computation of *Sum* aggregation for the query $Q(I; \theta)$ with the complain "the query value should be smaller" as a example. The algorithms for *Count* and *Mean* and other types of complains are similar.

First, the sum of the model output is $\sum_{x_i \in I} \frac{1}{1+e^{-\theta^\top x_i}}$. Since the complaint is for minimizing the sum, the corresponding gradient of the complaint is $Q' = -\sum_{i \in I} \frac{x_i}{e^{\theta^\top x_i} + 2 + e^{-\theta^\top x_i}}$. Directly sharing $Q'$ is not secure because doing so will leak $x_i$ in the nominator. However, we can let each party store their own gradients $\sum_{x_i \in I} \frac{x_i^A}{e^{\theta^\top x_i} + 2 + e^{-\theta^\top x_i}}$ and $\sum_{x_i \in I} \frac{x_i^B}{e^{\theta^\top x_i} + 2 + e^{-\theta^\top x_i}}$ and never share them with each other. On the other hand, $\theta^\top x_i$ in the denominator must be shared with the other party in plain text. Similar to the inference stage, computing $\theta^\top x_i$ is done by Party B sharing $(\theta^B)^\top x_i^B$ to Party A. We will discuss the security implication for this sharing in Section 4.3.

**Algorithm 1:** Protocol for computing $Hv$ in FEDRAIN

**Input:** $\theta^{\{A,B\}}$, $D^{\{A,B\}}$, $v^{\{A,B\}}$
**Output:** $Hv^{\{A,B\}}$
1 *Party A* Send $[\![R]\!]_A$ and $[\![RX^A v^A]\!]_A$
2 *Party B* Send $[\![Hv^B + \epsilon^B]\!]_A$ and $[\![X^B v^B]\!]_B$
3 *Party A* Send $[\![Hv^A + \epsilon^A]\!]_B$; then Decrypt & send $Hv^B + \epsilon^B$
4 *Party B* Decrypt and send $Hv^A + \epsilon^A$
5 *Party A* Derandomize $Hv^A + \epsilon^A$
6 *Party B* Derandomize $Hv^B + \epsilon^B$

---

**Algorithm 2:** Protocol for computing $Q'H^{-1}E$ in FEDRAIN

**Input:** $\theta^{\{A,B\}}$, $D^{\{A,B\}}$, $z^{\{A,B\}}$
**Output:** $Q'H^{-1}E$
1 *Party B* Send $(\theta^B)^\top x_i^B$ for $x_i^B \in D^B$
2 *Party A* Send $[\![y_i - h_\theta(x_i)]\!]_A$
3 *Party B* Send $(z^B)^\top [\![y_i - h_\theta(x_i)]\!]_A x_i^B$
4 *Party A* Compute
   $(z^A)^\top (y_i - h_\theta(x_i)) x_i^A + (z^B)^\top (y_i - h_\theta(x_i)) x_i^B$

---

*4.2.2 Compute $Q'H^{-1}$.* Directly compute the hessian matrix $H$ and then take the inverse to get $H^{-1}$ suffers from numerical instability issue [25]. Instead, computing $Q'H^{-1}$ can be treated as solving the unknown vector $z$ from the linear system $Hz = Q'$. We will use the *CG* [52] algorithm to solve $z$. Different from computing $Q'$, the *CG* process used to compute $Q'H^{-1}$ is more complex in that it needs to compute several auxiliary vectors and requires more communication between *Party A* and *Party B* because *CG* approximates $z$ in an iterative way. We will discuss a sketch on how to make *CG* work here and the detailed protocol can be found in our technical report[1].

On the data storage side, similar to $Q'$ that the value is separated into two parties, we did the same for the *CG* algorithm for maximizing the security. That is, we found that all the vectors inside *CG* are linearly separable across two parties which allows each party to store these vectors locally and never share with the other. For example, the residual vector $r = H\hat{z} - Q'$ of the approximated solution $\hat{z}$, which is used in the *CG* algorithm, is separated into two parties as $r^A$ and $r^B$ with each having the size $m^A$ and $m^B$. On the other hand, in terms of the computations, *CG* uses vector product $v^\top v$ and Hessian-Vector-Product (HvP) $Hv$ in each iteration. Among them, HvP requires some work to make it secure.

**Protocol for HvP** Hessian-vector-product takes a size $m$ vector $v$ and outputs another size $m$ vector $Hv$ by multiplying $v$ with the $m \times m$ hessian matrix. The protocol design of $Hv$ is based on block matrix notation. First, the hessian of logistic regression is

$$H = \sum_{x_i \in D} x_i x_i^\top h_\theta(x_i)(1 - h_\theta(x_i)) \qquad (6)$$

By using the matrix form $X$ where $X_i = x_i^\top$ and diagonal matrix $R_{ij} = h_\theta(x_i)(1 - h_\theta(x_i))$ to represent the hessian, we get $H = X^\top RX$. If we decomposite $X$ into $[X^A \; X^B]$, using block notation we get

---

[1] http://tiny.cc/fedrain-frog

---

**Table 2: Time Complexity on computing encrypted values of FEDRAIN vs. FROG.**

| | ENCRYPTED VALUES COMPUTATION | |
|---|---|---|
| Debugging | FEDRAIN | FROG |
| Influence | $O(pnm)$ | $O(nm)$ |
| Retraining | $O(qnm)$ | 0 |
| **Total** | $O(K \cdot pnm) + O(K \cdot qnm)$ | $O(K \cdot nm)$ |

$$H = \begin{bmatrix} (X^A)^\top RX^A & (X^A)^\top RX^B \\ (X^B)^\top RX^A & (X^B)^\top RX^B \end{bmatrix} \qquad (7)$$

After that, assuming $v$ is also separable into two parties: $v = [v^A \; v^b]$, $Hv$ will be:

$$\begin{aligned} Hv &= \begin{bmatrix} (X^A)^\top RX^A & (X^A)^\top RX^B \\ (X^B)^\top RX^A & (X^B)^\top RX^B \end{bmatrix} \begin{bmatrix} v^A \\ v^B \end{bmatrix} \\ &= \begin{bmatrix} (X^A)^\top RX^A v^A + (X^A)^\top RX^B v^B \\ (X^B)^\top RX^A v^A + (X^B)^\top RX^B v^B \end{bmatrix} \\ &= \begin{bmatrix} Hv^A \\ Hv^B \end{bmatrix} \end{aligned} \qquad (8)$$

Given the block matrix form of $Hv$, we can use the protocol in Algorithm 1 to compute it.

In step 2, *Party B* computes

$$[\![Hv^B]\!]_A = (X^B)^\top [\![RX^A v^A]\!]_A + (X^B)^\top [\![R]\!]_A X^B v^B \qquad (9)$$

and in step 3 *Party A* computes

$$[\![Hv^A]\!]_B = (X^A)^\top RX^A v^A + (X^A)^\top R[\![X^B v^B]\!]_B \qquad (10)$$

Also note that in line 2 and 3, two parties add random noises $\epsilon^A$ and $\epsilon^B$ to the encrypted data before sending them out to another party for decryption. This ensures the computation of Hvp remains secure without leaking any plain text data.

*4.2.3 Compute $Q'H^{-1}E$.* The gradient of one training sample for the logistic regression is $(y_i - \hat{y}_i)x_i$. Instead of sending the partial gradient $(y_i - \hat{y}_i)x_i^B$ from *Party B* to *Party A* in plaintext, which will leak the dataset $x_i^B$, we let *Party B* store this gradient locally and only send the computed influence score $Q'H^{-1}E$ to *Party A*. In detail, *Party B* first compute the partial influence score as $z^B(y_i - \hat{y}_i)x_i^B$ and then send it to *Party A* in plain text. The protocol is depicted in Algorithm 2.

*4.2.4 Time Complexity Analysis.* As mentioned in the preliminaries, computations happened on a encrypted values are 2 to 3 magnitudes slower than their unencrypted counterpart. Thus, in this section we report the time complexity for model training and debugging for the computations on the encrypted values only. The result is shown in Table 2.

**Training and retraining** *Party B* needs to multiply the encrypted residual by its training data $x_i^B$ and then sum up the result to get the gradient $\frac{\partial L(\theta)}{\partial \theta^B} = -\frac{1}{n} \sum_{i=1}^n [\![(y_i - h_\theta(x_i))]\!]_A x_i^B$ in each GD round. The complexity for doing $[\![y_i - h_\theta(x_i)]\!]_A x_i^B$ for $n$ training records is $O(nm)$. The summation takes another $O(n)$, so in total the complexity for a single GD round on the encrypted value is $O(nm)$. Assuming GD runs for $q$ rounds, the complexity for training and retraining is $O(qnm)$.

**Influence Calculation** During the computation of HvP, Party A needs to multiply an $n \times m$ vector $(X^A)^\top R$ with the encrypted $m \times 1$ vector $[\![X^B v^B]\!]_B$ as $(X^A)^\top R [\![X^B v^B]\!]_B$ in Equation (10), which takes $O(nm)$ complexity. Similarly, Party B needs to compute Equation (9) which also takes $O(nm)$. Since this process is running inside the CG algorithm in an iterative fashion, assuming the CG algorigthm iterate $p$ times, the complexity for CG is $O(pnm)$. Additionally, computing the influence score in Algorithm 2 Line 3 requires another $O(nm)$ computation similar to the training case. Overall, one debugging iteration takes $O(pnm)$ as the the complexity.

**Debugging** FEDRAIN debugs the training data in an iterative fashion where it repeatedly removes the top ranked points from the training set by influence score, and retrains the model. Assuming it deletes K records from the training set in total with each iteration deleting 1 record, the overall complexity for influence computation will be $O(K \cdot pnm)$ and the retraining complexity will be $O(K \cdot qnm)$. In total, it is $O(K \cdot pnm) + O(K \cdot qnm)$

### 4.3 Security Analysis

We provide theorems and proof sketches to analyze the security guarantee on the training and debugging protocol of FEDRAIN in this section.

#### 4.3.1 Training.

THEOREM 4.1. *The training process of* FEDRAIN *is secure under our security model in Definition 2.1 if the number of gradient descent round is less than $\frac{n \times m^B}{n - m^B}$, where $n$ is the number of samples in $D^B$ and $m^B$ is the number of features in $D^B$.*

As proved by the original author in [63], assuming the shape of the training dataset $D^B$ from Party B has size $n \times m^B$, the training protocol is secure if the number of GD round for the training stage is smaller than $\frac{n \times m^B}{n - m^B}$. As mentioned in Section 4.1, Party A knows $(\theta^B)^\top x_i^B$ for $x_i^B \in D^B$ from Party B's messages, which forms a nonlinear system with $n \times m^B + m^B$ unknowns ($m^B$ unknowns in $\theta$ and $m^B$ unknowns in $x_i^B$ with $n$ different $x_i^B$) and $n$ equations. There will be infinite number of solutions of $\{x_i^B\}_{i=1}^n$ due to the number of unknowns ($n \times m^B + m^B$) are larger than the number of equations ($n$). However, for each round, Party A will get $n$ new equations and $m^B$ new unknowns. After $r$ rounds, Party A will get $n \times m^B + r \times m^B$ unknowns and $n \times r$ equations. In order to keep $n \times r < n \times m^B + r \times m^B$, $r$ should be smaller than $\frac{n \times m^B}{n - m^B}$. □

#### 4.3.2 Debugging.

THEOREM 4.2. *The computation process of $Q'H^{-1}E$ for* FEDRAIN *is secure under our security model in Definition 2.1 if $m^B > 1$ and the number of debugging iteration is less than $\max\{\frac{n_D \times m^B}{n_D - m^B}, \frac{n_I \times m^B}{n_I - m^B}\}$, where $n_D$ is the number of samples in $D^B$ and $n_I$ is the number of samples in $I^B$.*

*Proof Sketch.* Throughout the debugging stage, there are several pieces of information from Party B that is known by Party A in plain text. They are, $(\theta^B)^\top x_i^B$ for $x_i^B \in I^B$ during the computation of $Q'$ in Section 4.2.1, the vector products $v^\top v$ during the computation of CG in Section 4.2.2 and the partial influence score $(z^B)^\top (y_i - h_\theta(x_i))x_i^B$ in Section 4.2.3. Similar to the training stage, in each debugging iteration Party A gets $n_I$ equations and $m^B + n_I \times m^B$ unknowns about $(\theta^B)^\top x_i^B$ for $x_i^B \in I^B$. This forbids the debugging iteration exceeding $\frac{n_I \times m^B}{n_I - m^B}$. On the other hand, knowing the partial influence score $(z^B)^\top (y_i - h_\theta(x_i))x_i^B$ by Party A is equivalent to knowing $(z^B)^\top x_i^B$ since Party A knows $(y_i - h_\theta(x_i))$. This leads to Party A learning $n_D$ equations and $n_D \times m^B$ unknowns from $(\theta^B)^\top x_i^B$ (Line 1 in Algorithm 2) and another $n_D$ equations and $n_D \times m^B$ unknowns from $(z^B)^\top x_i^B$ (Line 3 in Algorithm 2). This forbids the debugging iteration exceeding $\frac{n_D \times m^B}{n_D - m^B}$. On the contrary, knowing $v^\top v$ by Party A does not lead to security issue if $m^B > 1$, due to there are only 1 equation known by Party A with $m^B$ unknowns. □

## 5 FROG: EFFICIENT FEDERATED DEBUGGING

**FedRain Deficiencies.** FEDRAIN proposed in Section 4 shows a plausible way to enable SQL-based data debugging for federated learning. However, it has two major deficiencies which impedes it to be largely adopted in reality. One concern comes from security guarantee while the other is expensive time cost.

*Deficiency 1. Security Limitation.* FEDRAIN poses a very tight limit on the number of gradient descent iterations. For example, if party B contains 1000 rows and 10 features. To meet the security guarantee in section 2, the training protocol can only run a maximum of $\frac{n \times m}{n - m} \approx 10$ iterations. However, a normal logistic regression model typically needs far more iterations to converge in the real-world scenario.

*Deficiency 2. Expensive Time Cost.* As shown in Table 2, FEDRAIN also inflicts large overhead due to heavy encrypted gradient computation in debugging. In the system total running time, training (and retraining in debugging) is the major bottleneck of framework efficiency. This point can be explained by the fact that $q$ is usually a large number (e.g. 1000) for ML models to converge. In contrast, the number of CG iterations, $p$ is a relatively small number between 10 and 20. Therefore, optimizing encrypted gradient computation in training is the most crucial task in reducing total framework running time.

### 5.1 Linearly Separable Model Structure

To remedy the above two deficiencies, we propose FROG, an efficient and secure federated debugging framework.

In terms of expensive time cost in FEDRAIN, we cannot resolve this issue if sticking to the original exact logistic regression model structure. Each party needs to rely heavily on the information from the other party to compute gradients and influence values. A natural thought is to create a linearly separable model structure.

We represent the local logistic models stored in Party A and Party B by $f_1$ and $f_2$ as follows. For $(x^A, y) \sim D^A, x^B \sim D^B$,

$$f_1(x^A) = \frac{1}{1 + e^{-(\theta^A)^\top x^A}}, \quad f_2(x^B) = \frac{1}{1 + e^{-(\theta^B)^\top x^B}}$$

Based on these two local models, the linearly separable structure is expressed as $f(x) = f_1(x^A) + f_2(x^B)$. For this linearly separable structure, original negative log-likelihood loss function used in FedRain does not work anymore because the combined value is not a likelihood. Instead, we use the L2 loss function $\ell_i(\theta) = \frac{1}{2}(f(x_i) - y_i)^2$. The gradient of loss for training record $i$ then is:

$$\nabla \ell_i(\theta) = (f(x_i) - y_i) \nabla_\theta f(x_i)$$
$$= (f_1(x_i^A) + f_2(x_i^B) - y_i)(\nabla_{\theta^A} f_1(x_i^A) | \nabla_{\theta^B} f_2(x_i^B)) \quad (11)$$

With the linearly separable structure based on the L2 loss, we split the gradient computation into a local part, $\nabla_\theta f$ and a shared part, $(f - y)$. Both parties can perform a local gradient update with only the communication of $(f - y)$. Note that $(f - y)$ does not need to be encrypted. This implies we do not need to conduct expensive computation for gradients on encrypted values in training and thus we resolve the complexity issue.

To resolve the security issue, we add a *mask* to each party. Intuitively, for a linear system $Ax = b$, we mask the left part with a random number $c$ to transform the linear system into $cAx = b$. In this way, we protect $x$ from leakage. The masked linearly separable structure is expressed as $f = c_1 f_1 + c_2 f_2$. Then the gradient of loss becomes:

$$\nabla_\theta \ell_i(\theta) = (f(x_i) - y_i) \nabla_\theta f$$
$$= (c_1 f_1(x_i^A) + c_2 f_2(x_i^B) - y_i)(c_1 \nabla_{\theta^A} f_1(x_i^A) | c_2 \nabla_{\theta^B} f_2(x_i^B)) \quad (12)$$

where $c_1$ and $c_2$ are invisible to the other party. In this way, we resolve the security issue.

Note that if $c_1$ or $c_2$ is not a trainable variable, a fixed value will adversely affect the accuracy of the model. In our linearly separation design, we prove that we can have trainable $c_1, c_2$ without security concerns (see the proof in Section 5.3).

The gradient of $\ell_i(\theta, c_1, c_2)$ with respect to $c_1$ is

$$\nabla_{c_1} \ell_i(\theta, c_1, c_2) = (f(x_i) - y_i) f_1(x_i^A)$$

where $f_1(x_i^A)$ is in Party A and $f(x_i) - y_i$ is the same shared value as Equation (12). The same argument holds for $c_2$. We merge $c_1, c_2$ into $\theta^A, \theta^B$ for simplicity since their updating protocols are the same.

In general, the linearly separable structure enables both parties to update their own logistic models locally without exchanging encrypted raw data. The gradients of each party would then be naturally aggregated through model. This structure also benefits our debugging stage, which would be further explained in subsequent sections. Note that although the model under Frog is an approximation of the logistic regression mode, we show empirically in Section 6.3 that it can achieve quite close F1 scores in prediction to Rain [59], which has a centralized logistic regression model.

## 5.2 Working Mechanism of Frog: Training & Debugging

In this section, we discuss the protocols and working mechanisms in Frog. We first present how each local model is updated in the training stage. Then, we will show how the query gradients as well as the influence of each data point are calculated in our framework. For simplicity, we omit $c_1, c_2$ gradients updating details in following

**Algorithm 3:** Protocol for computing $\nabla \ell(\theta)$ in Frog

**Input:** $\theta^{\{A,B\}}$, $D^{\{A,B\}}$, $c_1$, $c_2$, $gd\_stop$
**Output:** $\nabla \ell(\theta)$ in equation (12)
1 **Party A** Send $c_1 f_1(x_i^A) - y_i$
2 **Party B** Send $c_2 f_2(x_i^B)$
3 **Party A** Compute gradient
   $(c_1 f_1(x_i^A) + c_2 f_2(x_i^B) - y_i) c_1 \nabla_{\theta^A} f_1(x_i^B)$
4 **Party B** Compute gradient
   $(c_1 f_1(x_i^A) + c_2 f_2(x_i^B) - y_i) c_2 \nabla_{\theta^B} f_2(x_i^B)$

---

**Algorithm 4:** Protocol for computing $\tilde{Q}'$ in Frog

**Input:** $\theta^{\{A,B\}}$, $D^{\{A,B\}}$, $c_1$, $c_2$
**Output:** $Q'$
1 **Party B** Send $Agg(c_1 f_1(x_i^A))$, $[\![Agg(c_2 \nabla \theta^B f_2(x_i^B))]\!]_B$
2 **Party A** Compute $Q'$; then send $[\![r_1 \cdot Q']\!]_B$

---

$$\begin{bmatrix} H^{AA} & H^{BA} \\ H^{AB} & H^{BB} \end{bmatrix}$$

**Figure 2: Symmetric properties of Hessian Matrix H.**

analysis since they can be trivially computed through the following protocols shown in training and debugging stages. The computation does not require any extra information to be shared between those two parties.

*5.2.1 Compute $\nabla \ell(\theta)$.* The information exchange protocols between two parties to compute the training gradient $\nabla \ell(\theta)$ are shown in Algorithm 3. The input symbol $gd\_stop$ in the protocol is a signal used to control whether both parties should continue the gradients updating iterations or not.

*5.2.2 Compute $Q'$.* The query gradient of model $f$ is as follows,

$$Q' = sgn(Agg(f(x_i)) \nabla_\theta Agg(f(x_i))$$
$$= sgn(Agg(c_1 f_1(x_i^A) + c_2 f_2(x_i^B)))$$
$$\cdot (\nabla_\theta Agg(c_1 f_1(x_i^A)) + \nabla_\theta Agg(c_2 f_2(x_i^B))),$$

where $Agg(\cdot)$ represents any aggregation function defined in section 2 that takes the *inference* data as input for the contained function. For example, $Agg(f(x_i))$ means compute the aggregation function on the output of $f$ which is computed on the inference data $I$. $sgn$ is the *sign* function, which would return 1 (0) if the expression inside is $\geq 0$ ($< 0$).

The communication protocols to calculate $Q'$ is shown in Algorithm 4. Note that in order to prevent party B to access the value of $Q'$, we randomize $Q'$ by $r_1$, denoted as $\tilde{Q}'$ in all following sections. Note this *scalar* randomization will not change our final influence ranking results.

*5.2.3 Compute $Q'H^{-1}$.* Different from FedRain, hessian matrix can be explicitly computed under Frog due to the linearly separable structure. Based on the symmetric property of hessian matrix, the hessian matrix can be divided to four parts as shown in Figure 2.

**Algorithm 5:** Protocol for computing $\tilde{Q}'H^{-1}$ in Frog

**Input:** $\theta^{\{A,B\}}$, $D^{\{A,B\}}$, $c_1, c_2$
**Output:** $\tilde{Q}'H^{-1}$
1 **Party B** Send $[\![c_2 \nabla_{\theta_B} f_2(x_i^B) + \epsilon]\!]_B$
2 **Party A** Compute $[\![\hat{H}^{AB}]\!]_B$; then send $[\![\hat{H}^{AB}]\!]_B, \sum_j c_1 \nabla_{\theta_A} f_1(x_i^A), H^{AA}$
3 **Party B** Decrypt and derandomize $[\![\hat{H}^{AB}]\!]_B$; then construct $H$
4 **Party B** Compute $\tilde{Q}'H^{-1}$ using CG Algorithm

---

**Algorithm 6:** Protocol for computing $\tilde{Q}'H^{-1}E$

**Input:** $\theta^{\{A,B\}}$, $D^{\{A,B\}}$, $c_1, c_2$
**Output:** $Q'H^{-1}E$
1 **Party B** Send $(\tilde{Q}'H^{-1})^{m^A}$
2 **Party A** Send $(\tilde{Q}'H^{-1})^{m^A}(E^A)$
3 **Party B** Send $(\tilde{Q}'H^{-1})^{m^B}(E^B)$
4 **Party A** Compute $\tilde{Q}'H^{-1}E$
5 **Party B** Compute $\tilde{Q}'H^{-1}E$

---

The formulas to calculate for each part is as follows:

$$H^{AA} = \nabla_{\theta^A}(\nabla_{\theta^A} \ell(\theta))$$
$$= \nabla_{\theta^A}\left[\sum(f-y)c_1 \nabla_{\theta^A} f_1\right]$$
$$= \sum(f-y)c_1 \nabla^2_{\theta^A} f_1 + (c_1 \nabla_{\theta^A} f_1)^2$$
$$H^{AB} = \nabla_{\theta^B}(\nabla_{\theta^A} \ell(\theta))$$
$$= \sum c_2 \nabla_{\theta^B} f_2 \cdot c_1 \nabla_{\theta^A} f_1$$

Note that $H^{BB}$ has a similar format with $H^{AA}$, and $H^{BA}$ is just the transpose of $H^{AB}$, so they are omitted here for simplicity. The calculations of $H^{AA}$ and $H^{BB}$ can be completed in each party *locally*, since they doesn't require any information from the other party.

The communication protocol to calculate $\tilde{Q}'H^{-1}$ is shown in Algorithm 5, where $\hat{H}^{AB}$ is the $H^{AB}$ computed with randomized $c_2 \nabla_{\theta^B} f_2$. Note that in the whole $\tilde{Q}'H^{-1}$ computing process, the *CG* is only used *once* and computed locally in *Party B*.

#### 5.2.4 Compute $Q'H^{-1}E$.
The protocol of computing the final influence function is displayed in Algorithm 6. From Algorithm 5, we can know that $\tilde{Q}'H^{-1}$ locates in *Party B*. To enable *Party A* to compute its own influence, *Party B* only needs to send a vector $(\tilde{Q}'H^{-1})^{m^A}$, which represents the part belonging to *Party A* in $(\tilde{Q}'H^{-1})$ rather than the whole $\tilde{Q}'H^{-1}$. This is because Frog is linearly separable, similar for the meaning of $(\tilde{Q}'H^{-1})^{m^B}$. After this step, both parties can calculate their own influence with their element-wise gradients $E^A$, $E^B$ locally. At last, each party would send their influence results to the other to compute the final influence rankings based on the additive property of an influence function. Thus, each party would know which training data points to delete in the end.

#### 5.2.5 Time Complexity Analysis.
Computation on encrypted values is the major bottleneck of our debugging frameworks as mentioned in section 3 and 4.2.4. In this section, we analyze the debugging time complexity of Frog. The results are summarized in Table 2.

**Influence Calculation.** The largest computation cost comes from $[\![\hat{H}^{AB}]\!]_B$, whose complexity is $O(nm)$. Supposing we have $K$ total rounds of debugging, the cost would be $O(K \cdot nm)$.

**Retraining.** The retraining protocol in debugging is the same as training shown in Algorithm 3. There is no encrypted values in training, therefore the computation cost on encrypted values is 0, no matter how many times the debugging runs.

### 5.3 Security Analysis

In this section, we analyze the security of training and debugging protocols of Frog.

#### 5.3.1 Training.

**Theorem 5.1.** *The training process is secure under our security model defined in Definition 2.1.*

*Proof Sketch.* Training protocols shown in Algorithm 3 leaks no information of raw data to the other party. In training, party A would receive a vector of $c_2 f_2$ in every stochastic gradient descent iteration. As mentioned in section 5.1, $c_1, c_2$ are trainable variables. It implies we update $c_1$ with zero additional communication from party B, and similar for $c_2$. Specifically, there would be unique $(n + 1)$ unknown variables for each iteration while there are only $n$ equations.

Note that $c_2$ and $f_2$ changes with each iteration, which implies $c_2$ and $f_2$ are unsolvable.

To this end, party A cannot decode $\theta^B$ and $x^B$. Similar logic applies party B in the training stage communication.

#### 5.3.2 Debugging.

**Theorem 5.2.** *The debugging process is secure under our security model defined in Definition 2.1, assuming $n > m$.*

*Proof Sketch.* First, let's check whether party A or B can decipher any raw data, gradients or $\theta$ of each other during the Q gradient computation process. As shown in Algorithm 4, each equation containing plain-text or decipherable information from the other party is randomized by a **unique** variable (e.g. $r_1, r_2, r_3$), which makes them unsolvable. This implies that for later constituted underdetermined system of equations, those equations for computing $Q'$ in this process cannot be counted towards them anymore to compose a possible solvable equation system.

In Algorithm 5, the decipherable information and plain-text of party A received by party B are $\hat{H}^{AB}$, $H^{AA}$ and $\sum_j g_j$. The number of equations can be constructed by party B is $m^A \times (m + 1)$, while the number of unknowns w.r.t $\theta^A, x_i^A, y_i$ is $m^A \times (n+1) + n$. As long as $n > m$, namely, the number of data samples is larger than the number of features, the equation system is unsolvable. For party A, the plain information received about party B is $\tilde{Q}'H^{-1}$. It's an approximate value, which means no certain information can be obtained from this value.

As for the influence calculation shown in 6, the passed information $(\tilde{Q}'H^{-1})^{m^B}(E^B)$ from party A to party B can allow B to construct more $n$ equations. After this process, the total accumulated number of equations in party B is $m^A \times (m+1) + n$. However, it's still less than the number of unknown variables, $m^A \times (n+1) + n$, as long as $n > m$.

## 5.4 Overall Comparison: Frog vs. FedRain

In this subsection, we give an overall analytical comparison between FedRain and Frog from three perspectives: efficiency, security, and accuracy.

*5.4.1 Efficiency.* We summarize the time complexity of performing computation on encrypted values in the debugging stage of both frameworks in Table 2. Recall that in the table, $p$ represents the number of CG iterations in one debugging round, and $q$ is the number of gradient descent epochs of one debugging round. The total time complexity is calculated by assuming we debug for $K$ rounds. As illustrated in Algorithm 3, there is no encryption cost at all in Frog during retraining (training). Thus the time complexity of training on computing encrypted values is 0. Besides, the CG computation of Frog is conducted locally as explained in Algorithm 5 while FedRain needs to do CG computation several times.

*5.4.2 Security.* The assumption of security guarantee made by FedRain is stronger than Frog. To ensure security, the FedRain debugging protocol can run no more than $\frac{n \times m}{n-m}$ iterations. In contrast, as long as the number of training examples is larger than the number of features, Frog can guarantee security under the security model defined in Definition 2.1.

*5.4.3 Accuracy.* The model used in Frog is an approximation to the exact logistic regression model. In comparison, FedRain is based on the exact logistic regression model. One may be tempted to think that FedRain should be more accurate than Frog. However, to ensure security, FedRain has to pose a very tight limit on the number of gradient descent iterations. In many situations, this number is too small for FedRain to converge to an accurate model, thus Frog is actually more accurate than FedRain.

## 6 EXPERIMENTS

Our experiments study the extent and how Frog improves upon FedRain in terms of runtime and debugging quality. Section 6.2 provides a break down of the computational and communication costs between the two methods and their sensitivity to dataset sizes and debugging parameters, and Section 6.3 studies the debugging quality as compared to FedRain and loss-based approaches. Additionally, Section 6.4 presents a case study that uses Frog to improve the fairness on the Adult dataset.

### 6.1 Experiment Settings

Our experiment are setup as follows:
**Dataset:** We used three datasets: *Diabetes* [19] contains 442 rows, 10 numeric features and 1 continuous prediction target with domain $25-346$ that we threshold at the median ($> 140.5$) into a binary label. *BreastCancer* [18] contains 569 rows, 30 numerical features, and binary label. *Adult* [18] contains 32561 rows, a mix of 13 categorical and numeric features, and binary label.

We split each dataset so 80% is used for training, 10% for inference and SQL querying, and the rest as a hold-out to report the model accuracy as the result of debugging. We vertically split each dataset into two partitions, where the first half of attributes are on *Party A* and the rest on party *Party B*, and the ID is on both parties. Finally, we introduce label-dependent training data errors by flipping a random subset of '1' labels to '0' in the dataset.

Table 3: Time cost comparison between Federated LC Rain and Federated Logistic Rain.

|  |  | Frog | FedRain | Relative |
| --- | --- | --- | --- | --- |
| Training | Compute | 9.15s | 144.3s | 15× |
|  | Network | 0.09s | 13.2s | 146× |
| Debugging Influence | Compute | 0.62s | 17.7s | 28× |
|  | Network | 0.00075s | 0.7s | 933× |
| Debugging Retraining | Compute | 0.924s | 54.4s | 58× |
|  | Network | 0.008s | 6.5s | 812× |

**Approaches:** We compare the Rain [59] (which is not secure), the loss-based approach (Loss), FedRain, and Frog. We do not compare with Influence Function since it is subsumed by Rain.

**Evaluation Metrics** We measure end-to-end runtimes, as well as a break-down into computate and communication runtimes in the training and debugging steps. For debugging quality, we report the recall@k curve (Figure 4), which is the percentage of correctly identified training records as a function of the total number of deleted rows (k). We also report the model F1 scores before and after debugging on the hold-out dataset (Table 4).

**Implementation** Our implementation is in Python. We use PHE [16] for Paillier encryption with GMP [21] and gmpy2 [26] acceleration. We simulate a two party federated learning environment by running *Party A* and *Party B* in isolated 16G RAM and 8 core E7-4839 CPUs docker containers installed with Ubuntu 20.04 and use the docker bridge network (9 Gbps) for communication. The code is hosted on Github[2].

### 6.2 Evaluation of System Efficiency

We first study a runtime break down of the two federated debugging algorithms, and then study their parameter sensitivity.

*6.2.1 Runtime Breakdown.* Although we compared Frog and FedRain's time complexities analytically, we now compare their runtimes on the Diabetes dataset for each step of training and debugging, and for both computation and network transmission. We run training for 1000 gradient descent iterations. During debugging, we delete 10 training points (influence) and then retrain for 100 gradient descent iterations. The inference query computes the probability of diabetes for each gender, where Predictions is the materialized predictions table:

```
SELECT AVG(P.Label_prob())
FROM Predictions P ⋈_{ID} Diabettes Group BY GENDER
```

Table 3 reports the breakdown and confirms our analysis in Section 5.4. Frog reduces compute during (re)training for two reasons. First, it avoids computing over encrypted data. Second, the data sent from *Party B* to *Party A* is an unencrypted scalar, as opposed to two $n \times 1$ vectors per gradient iteration for FedRain. Frog speeds accelerates the CG algorithm when estimating the influence scores because it only needs to send and encrypt an $n \times 1$ vector once in total, as opposed once per CG iteration (10~20 times).

*6.2.2 Sensitivity to Parameters.* We now vary the dataset sizes (n), the number of features in the datasets (m), number of training records to delete during debugging (K), and report the end-to-end runtime sensitivities of the two approaches. We use the Diabetes dataset, and replicate its rows and attributes to achieve the desired

---
[2]https://github.com/sfu-db/FedRain-and-Frog

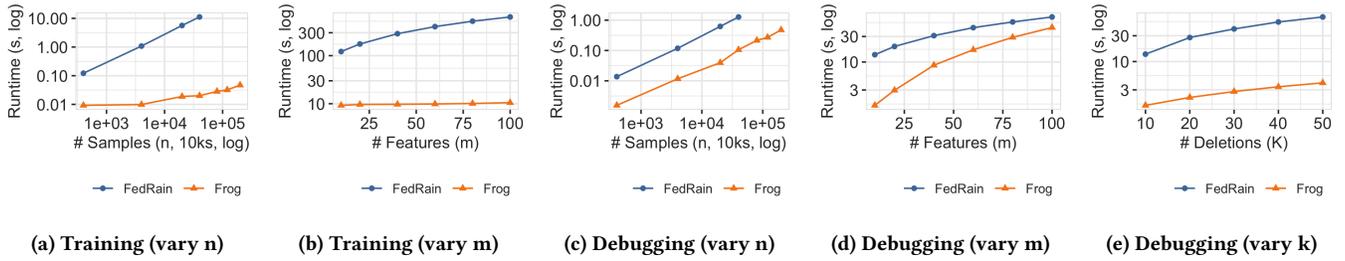

(a) Training (vary n)   (b) Training (vary m)   (c) Debugging (vary n)   (d) Debugging (vary m)   (e) Debugging (vary k)

Figure 3: Varying dataset size (n), number of features (m), and deletion size (K). A run was stopped if the runtime exceeded 3.5 hours.

configuration. The training errors and SQL query are the same as the previous experiment.

The results in Figure 3 are consistent with our complexity analysis in Section 5.4. We report the training (subfigures a, b) and debugging runtimes (c, d, e) in log scale as we vary each parameter (note that $K$ is only applicable to debugging). FedRain increases linearly for m and n during training, however Frog remains nearly constant because it does not require any encryption. In contrast, both methods scale linearly in terms of n and m during debugging because both have time complexity $O(nm)$ (though Frog is over an order of magnitude faster). Similarly, both methods scale linearly with $K$, but the slope for Frog is lower. Overall, we find that the major bottleneck is due to encryption before communication, which potentially can be reduced by custom hardware.

## 6.3 Data Debugging Quality

We now report the data debugging quality of Frog and FedRain, along with the Loss baseline. We also include Rain results running on a single machine setting (thus does not require the secure protocols). We use the Diabetes and Breast Cancer datasets, and introduce low (30% of records) and high (50%) corruption rates respectively. We execute a COUNT(*) query where the complaint specifies that the output should be the ground truth query result.

Overall, we find that the training records Frog returns is comparable to Rain. Note that Frog uses the linearly separable variant of the logistic regression model, and that FedRain is limited to a finite number of gradient iterations during training in order to preserve its security guarantees.

### 6.3.1 Recall@K.
Section 6.3 reports the Recall@K curves for the four approaches, and the gray line denotes the upper bound where every removed record was correctly identifies as an error.

Figure 4 shows the recall curve for different datasets with low (30%) and high (50%) corruption rates. Frog is comparable to Rain [59] in terms of recall across the datasets and corruptions, while Loss and FedRain performs poorly. Loss has a low recall curve because minimizing training loss during data debugging is, in general, independent of resolving user complaints. FedRain performs poorly because the model did not train to convergence due to its limited number of training gradient iterations in order to ensure the federated security guarantee. For example, in the Diabetes dataset, FedRain only runs about 10 gradient descent steps. For this reason, its influence gradient estimates are also inaccurate.

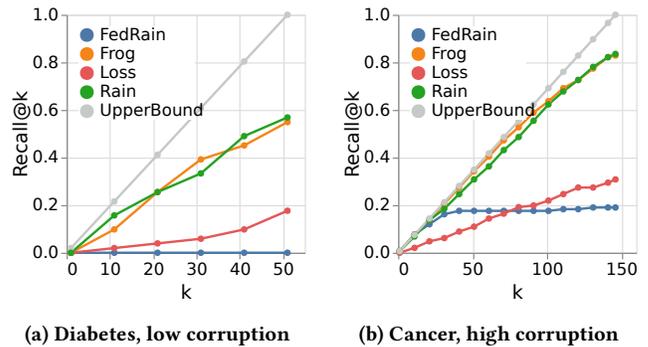

(a) Diabetes, low corruption   (b) Cancer, high corruption

Figure 4: Recall@k curve with varying corruption rates on Diabetes and BreastCancer. The K is equal to the total number of corrupted data points. The Recall@k performance of Frog is on par to Rain. Loss performs badly. FedRain is not fully trained due to its security guarantee in section 4.3 which therefore results in bad performance.

Table 4: Model F1 score before and after debugging; the clean data row is for reference.

| Model | Diabetes (30%) | | BreastCancer (50%) | |
| --- | --- | --- | --- | --- |
| | Before Debug | After Debug | Before Debug | After Debug |
| Rain | 0.72 | 0.81 | 0.58 | 0.84 |
| Loss | 0.73 | 0.74 | 0.56 | 0.64 |
| FedRain | 0.55 | 0.64 | 0.17 | 0.17 |
| Frog | 0.73 | **0.82** | 0.56 | **0.83** |
| Clean Data | - | 0.85 | - | 0.87 |

### 6.3.2 Model Accuracy.
Table 4 report the model F1 scores before and after debugging. The accuracy of Frog achieves the highest accuracy of the secure approaches and achieves comparable accuracy to Rain. In addition, it is close to the model accuracy trained on the clean dataset. In contrast Loss and FedRain marginally improve the model accuracy.

## 6.4 Case Study

We now illustrate how Frog can be useful to debug and address a gender bias issue in the context of a high-tech company that wants to collaborate with an HR agency to determine employee salaries. However, several female employees find that their salaries are much lower than male coworkers who are at the same position

Table 5: High salary prediction discrepency between female and male employees reduces after debugging with FROG.

|  | Discrepancy Before Debugging | Discrepancy After Debugging |
|---|---|---|
| Female | 20.89% | 48.91% |
| Male | 79.11% | 51.09% |

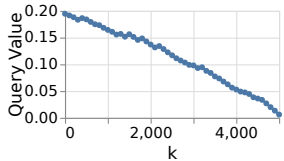

(a) The query value during debugging.

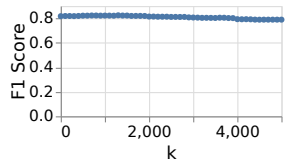

(b) Model F1 score changes on testing data when debugging.

Figure 5: Case Study Results

and have similar work performance. After hearing those complaints, the company issues the following query to check the average salary bands (high salary vs. low salary) between the two genders, where Predictions is the materialized predictions from the secure co-inference step:

```
SELECT avg(SALARY) FROM Predictions P ⋈_ID Adult
WHERE P.Label = 'High Salary'
GROUP BY GENDER
```

They are surprised to find a large salary discrepancy between female and male employees. There are several existing methods to address such biases, including data pre-processing by flipping labels of data points [28, 65], leveraging adversarial learning and regularization techniques to generate a fair model [66] and [38, 56]. Unfortunately, these methods either do not have secure federated protocols, or would require prohibitively high overheads due to encryption or complex model loss structures. In contrast, FROG can be used to debug this discrepancy.

To simulate this scenario, we use the UCI Adult [7] dataset with the simulated label errors. We issue the above query across the two parties, and submit a complaint that the discrepancy between the percentage of high salary employees of the two classes should be 0. Table 5 reports the percentages before and after debugging. We can see that percentage of high salary predictions between the genders changed from 20.89% vs. 79.11% before debugging to 48.91% vs. 51.09% after debugging, which is close to equal.

One worry may be that data debugging in this way addresses the discrepancy, but otherwise degrades the model accuracy. Figure 5a reports the discrepancy as a function of the number of deleted training records $K$, as well as the $F_1$ score on the hold-out data set. Figure 5a shows that the complaint is steadily and consistently resolved as we identify and remove the identified training errors. In contrast, Figure 5b shows that the $F_1$ score is almost constant throughout the data debugging process.

## 7 RELATED WORK

Complaint-based federated data debugging is related to data cleaning for ML, federated learning, ML pipeline debugging, and SQL explanation.

**Data Cleaning for ML.** ML for data cleaning [24, 37, 47] and data cleaning for ML [33, 34, 61] are active research topics, where the former studies how to use ML techniques to clean data and the latter explores how to clean data towards an accurate downstream ML model. Recent surveys have argued for the importance of using downstream applications for data debugging [42, 48], and this work extends prior approaches that leverage downstream complaints [32, 59] to a federated setting.

**Federated Learning.** Data privacy is a major issue in organizations, governments, and societies, as evidenced by numerous government regulations [1, 5, 12]. As a result, federated learning [10] is increasingly relied upon in industries such as finance, medicine, and transportation [2, 3, 11, 41, 54, 62, 64].

Federated learning methods have been developed for cases where the training data is applied horizontally (HFL) [14, 53, 55], or vertically (VFL) [15, 23, 55, 60] partitioned across the parties. There is limited work on debugging training data for federated learning. Chen et al. [14] studied how to handle label quality disparity in federated learning. They designed an algorithm to aggregate client models' updates based on a data quality measure of each client under HFL. In contrast, our work focuses on the more complex VFL setting and uses SQL-based complaints to fix data.

**ML Pipeline Debugging.** Data errors are a major issue in modern machine learning pipelines [8, 46]. Diagnostic debuggers as Data X-ray [57] help detect some data errors based on their common properties. Data validation and model assertions are commonly applied before model training and deployment [9, 20, 30]. The use of downstream model semantics or analytics for data debugging is relatively new, and recent works have leveraged model convexity [34], robustness [48], and queries [59].

**SQL Explanation.** SQL explanation uses query result complaints to detecting input data errors [4, 29, 40, 49, 50, 58]. This is a powerful concept since a user only needs to specify a high-level complaint and then the system will help the user to trace the complaint back to the corresponding data errors. FROG extends this debugging model to identify *training* rather than query input errors, and to a federated setting.

## 8 CONCLUSION

In this paper, we studied how to enable SQL-based training data debugging for federated learning. This is the first study on this important topic. We focused on logistic regression and two-party vertical federated learning, and formally defined our problem. We successfully extended Rain to a federated learning setting and call our framework FEDRAIN. We proved a security guarantee for FEDRAIN and analyzed its time complexity. After that, we identified the limitations of FEDRAIN, and proposed FROG, a novel federated debugging framework. A novel idea proposed in FROG is to modify the logistic regression model structure to make it more tailored for federated learning. Both theoretical analysis and experimental results showed that FROG is more secure, more accurate, and more efficient than FEDRAIN. In the end, we did a case study to demonstrate the effectiveness of FROG to resolve a real SQL-based complaint.

# A CG

In this section we describe how to adapt the CG algorithm (Algorithm 7) to the two party Federated Learning setting. As the premises, we assume $Q'$ is already computed and distributed into two parties as $Q'^A$ and $Q'^B$ as described in Section 4.2.1. For initializing $r_0$ in Line 1, the two parties first compute $Hz_0^{\{A,B\}}$ using the HvP protocol Algorithm 5. After that, they can compute $r_0^{\{A,B\}}$ locally. For *Party A* to decide whether $r_0$ is sufficiently small in Line 4 and Line 11, we use the L2 norm of the vector for $r_0$ and $r_k$. In detail, the L2 norm of a vector $v$ is $||v||_2 = v^\top v$ which can be separated into two parts as $(v^A)^\top v^A$ and $(v^B)^\top v^B$. So first *Party A* and *Party B* can compute each part separately and then *Party B* send $(v^B)^\top v^B$ to *Party A* to compute $||v||_2$. We describe this protocol in Algorithm 8. Note that this protocol is repeatedly used in Line 4 (for computing $r_0^\top r_0$), Line 6 (for computing $r_{k+1}^\top r_{k+1}$ and $p_k^\top H p_k$), Line 11 (for computing $r_k^\top r_k$) and Line 12 (for computing $r_k^\top r_k$ and $r_{k+1}^\top r_{k+1}$). On receiving the vector products in Line 6 and Line 12, *Party A* can then compute $\alpha_k$ and $\beta_k$ and send them to *Party B* for *Party B* to update $z_{k+1}$ (Line 7), $r_{k+1}$ (Line 8) and $p_{k+1}$ (Line 13).

---

**Algorithm 7:** Conjugate Gradient Algorithm

**Input:** $Q'^{\{A,B\}}$, $\theta^{\{A,B\}}$, $x_i^{\{A,B\}}$ for $x_i^{\{A,B\}} \in D$, $y_i$ for $y_i \in D$
**Output:** The result of $Q'H^{-1}$

1. initialize $r_0 := Q' - Hz_0$, $p_0 := r_0$, $k := 0$;
2. **if** $r_0$ *is sufficiently small* **then**
3.     **return** $z_0$
4. **end**
5. **repeat**
6.     $\alpha_k := \dfrac{r_k^\top r_k}{p_k^\top H p_k}$
7.     $z_{k+1} := z_k + \alpha_k p_k$
8.     $r_{k+1} := r_k - \alpha_k H p_k$
9.     **if** $r_{k+1}$ *is sufficiently small* **then**
10.         exit loop
11.     **end**
12.     $\beta_k := \dfrac{r_{k+1}^\top r_{k+1}}{r_k^\top r_k}$
13.     $p_{k+1} := r_{k+1} + \beta_k p_k$
14.     $k := k + 1$
15. **return** $z_{k+1}$

---

**Algorithm 8:** Protocol for computing $v^\top v$

**Input:** $v^{\{A,B\}}$
**Output:** $v^\top v$

1. *Party B* Send $(v^B)^\top v^B$
2. *Party A* Compute $v^\top v = (v^A)^\top v^A + (v^B)^\top v^B$